\documentclass{article}

\makeatletter
\def\input@path{{venue/iclr2027/}}
\makeatother
\usepackage{iclr2027_conference,times}

\usepackage{amsmath,amsfonts,bm}

\def\eqref#1{equation~\ref{#1}}

\def\1{\bm{1}}

\DeclareMathAlphabet{\mathsfit}{\encodingdefault}{\sfdefault}{m}{sl}
\SetMathAlphabet{\mathsfit}{bold}{\encodingdefault}{\sfdefault}{bx}{n}

\usepackage{amsmath}
\usepackage{amssymb}
\usepackage{booktabs}
\usepackage{graphicx,float}
\usepackage{hyperref}
\usepackage{microtype}
\usepackage{url}
\usepackage{xcolor}

\graphicspath{{figures/}{./}}

\newif\ifdraftmode
\draftmodefalse

\title{GREP: Grounding Generated Room Evaluation in Principles}

\author{Mohan Cao$^{1,2}$, Yongsen Mao$^{3}$, Chucheng Xiang$^{2,4}$, Zhongyuan Liu$^{2}$, \\
\textbf{Zhihang Zhong$^{5}$, Haiping Wang$^{3}$\thanks{Corresponding authors.}\hspace{0.4em} \& Yuan Liu$^{3}$\footnotemark[1]} \\
$^{1}$University of Washington \quad $^{2}$Tencent AI Lab \\
$^{3}$The Hong Kong University of Science and Technology \\
$^{4}$University of Science and Technology of China \quad $^{5}$Shanghai Jiao Tong University}

\iclrfinalcopy

\newcommand{\frozenassets}{FrozenAssets\textsuperscript{*}}

\begin{document}

\maketitle
\lhead{}

\vspace{-25pt}
\begin{figure}[H]
\centering
\includegraphics[width=\linewidth,height=0.64\textheight,keepaspectratio]{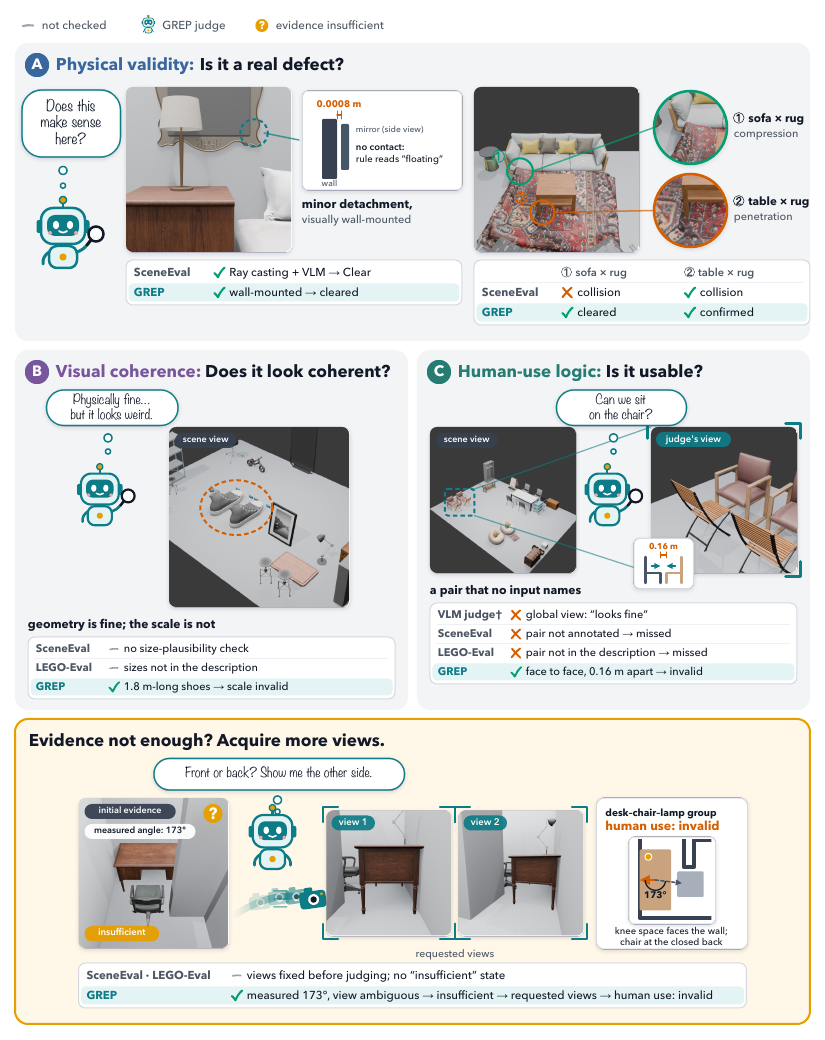}
\vspace{-15pt}
\caption{GREP is a scoring framework for generated scenes, grounded in three progressive \textit{principles}: (A) physical validity, (B) visual coherence, and (C) human-use logic. In contrast, existing metrics rely on narrow criteria or hard metrics, failing to assess the scene quality accurately. Moreover, GREP lets the evaluation agent \textit{autonomously} identify issues, gather evidence, and reach comprehensive judgments. Check and cross marks indicate whether each evaluator's output matches the actual state of the scene, not whether the scene passes. In (A), a pure contact rule would read the mirror 0.0008\,m from the wall as floating; both VLM-assisted evaluators clear it, whereas only GREP separates the intended sofa--rug contact from the table--rug penetration. $^\dagger$A holistic VLM judge given only the global view.}
\label{fig:teaser}
\end{figure}
\clearpage
\begin{abstract}
Recent LLM-centric advances can generate attractive furnished rooms by selecting and arranging assets. Yet evaluation remains limited to specific perspectives or predefined questions, failing to reflect overall scene quality. To address this pressing need, we introduce \textbf{GREP}. Its core is a focus shift from defining specific questions to establishing a three-tier evaluation philosophy: basic \textit{physical validity} checks whether a scene obeys physical constraints; \textit{visual coherence} concerns compatibility in object scale, style, and composition; and, at the highest level, \textit{human-use logic} requires object arrangements to align with everyday human habits.
Guided by these principles, an evaluation agent autonomously and adaptively identifies issues from scene context, gathers evidence, and produces scores and reports. This supports scene-specific issue discovery and reliable judgments.
Experiments show that (1) GREP achieves 87\% agreement with human
majority preferences, substantially outperforming baseline evaluators.
(2) GREP reports improve scene quality through refinement, whereas
score-only and no-feedback baselines show slight declines. (3) GREP suggests that as planning ability improves, recent models such as GPT-5.6 Sol can move from relying on pipelines and harnesses to challenging their guidance and constraints, but still lack the ability to resolve the resulting conflicts. Stronger models are therefore expected not only to plan well, but also to adapt to external guidance and constraints more flexibly. Project page: \href{https://han-808.github.io/GREP/}{\mbox{\nolinkurl{han-808.github.io/GREP}}}.
\end{abstract}

\section{Introduction}
\label{sec:introduction}

Given a room description, recent generation systems use language models \citep{layoutgpt2023,ran2025direct}, specialized pipelines \citep{holodeck2024,layoutvlm2024}, or harnessed agents \citep{yang2025sceneweaver} to select three-dimensional (3D) assets and arrange them into a furnished scene. The output specifies each object's size, position, and orientation. Evaluation must answer a broader question: does this arrangement make sense as a room?

As generated scenes grow more diverse, evaluation has not kept pace.
Existing methods assess particular aspects of scene quality, including
instruction-derived constraints \citep{legoeval2026},
text fidelity and physical plausibility \citep{tam2026sceneeval},
or activity-specific functional feasibility \citep{maillard2026sceneteract}.
Scene-specific problems outside these evaluation scopes may therefore
be overlooked, limiting assessment of the scene as a whole from a human
perspective (Fig.~\ref{fig:teaser}).

We introduce \textbf{GREP} (\textbf{G}rounding generated \textbf{R}oom \textbf{E}valuation in \textbf{P}rinciples), an evaluation framework that addresses this gap. Our core insight lies in {\textit{establishing a principled three-tier evaluation philosophy, rather than in defining specific checks.}} First, at the most basic level, a scene must follow \emph{physical validity}: objects must have adequate support and avoid unintended collisions or interpenetration. Next, \emph{visual coherence} requires objects to have compatible styles and sensible relative scales. Finally, at the highest level, real-world usability requires obeying \emph{human-use logic}, spanning open problems; for example, facing chairs need room for people to sit, and a desk chair should face the desk's usable side, as illustrated in Fig.~\ref{fig:teaser}.

This philosophy provides consistent principles, while the specific questions to ask vary from scene to scene. Our second contribution puts it into practice through {\textit{scene-specific inspection with adaptive evidence collection}}. An evaluation agent uses room context to identify objects and relations that warrant inspection, including those absent from the input description. It then autonomously uses geometric measurement, existing-view selection, and additional rendering to gather the evidence each check requires.
Finally, GREP aggregates these judgments into metric and scene scores, but its output goes beyond numbers. A detailed report identifies affected objects, explains the judgments, and records supporting evidence and unresolved gaps.
Encouragingly, we further show that \textit{such diagnostic reports help generators largely improve scene quality} by pinpointing specific problems and guiding targeted revisions.

Our experiments have two parts. First, we use GREP to evaluate current scene generation approaches. The results suggest that, as planning improves, models such as GPT-5.6 Sol can move from relying on pipelines and harnesses to challenging their guidance and constraints. Yet they still struggle to resolve the resulting conflicts, compromising scene quality. This calls for future models and systems that combine strong planning with the flexibility to adapt to external guidance and constraints. Second, we assess GREP itself. It achieves 87\% agreement with human preferences, substantially outperforming baseline metrics. In a five-scene study with GPT-4.1, report-guided refinement
increases the mean score from a shared initial value of 0.585
to 0.676 after three revision rounds, compared with 0.543
without feedback, using the refinement scoring protocol
described in Appendix~\ref{app:feedback-protocol}.

\section{Related Work}
\label{sec:related-work}

\textit{LLM-driven Scene Generation.} Earlier learned generators synthesize layouts from data priors with autoregressive or diffusion models \citep{paschalidou2021atiss,tang2024diffuscene,lin2024instructscene}, and PhyScene adds physical interactability \citep{physcene2024}. Language-guided scene generation systems differ in representation and control. LayoutGPT and DirectLayout use language-model reasoning to predict numerical placements \citep{layoutgpt2023,ran2025direct}; Holodeck and LayoutVLM incorporate spatial constraints or optimization into scene construction \citep{holodeck2024,layoutvlm2024}; SceneWeaver coordinates scene-generation tools through an iterative agent \citep{yang2025sceneweaver}; and recent layout agents and co-optimization methods extend these designs \citep{wan2026nala,xiang2026co}. This diversity calls for evaluation independent of the generation mechanism. GREP accepts scenes, assets, transforms, and spatial context through a common interface.

\textit{Structured Scene Evaluation.} Structured evaluators decompose tasks into explicit requirements or verification programs. Davidsonian Scene Graph (DSG) uses atomic questions and dependencies to assess text-to-image faithfulness \citep{cho2024dsg}; weak-verifier approaches generate task-specific programs for spatial layouts \citep{zhang2026weakverifiers}; and SceneEval, LEGO-Eval, and SceneTeract verify physical, instructional, or activity-specific properties in generated 3D scenes \citep{tam2026sceneeval,legoeval2026,maillard2026sceneteract}. These methods provide valuable checks, but their check vocabularies are primarily fixed by input requirements, annotations, or supplied activities. Model-based judges instead score outputs directly, from reference-free and question-answering metrics \citep{hessel2021clipscore,hu2023tifa} to LLM and VLM judges \citep{zheng2023judging,lee2024prometheusvision}; GREP also uses a VLM, but conditions each judgment on a scene-specific check and its retained evidence. GREP organizes evaluation around physical validity, visual coherence, and human-use logic. This three-tier philosophy guides scene-specific checks, including relations that emerge from the generated arrangement.

\textit{Evidence-Grounded Visual Reasoning.} Task-directed observation selection reveals information missing from a fixed view. AffordBot, Agentic Affordance Reasoning (A3R), and Chain-of-View (CoV) acquire complementary visual or geometric evidence for affordance and embodied reasoning tasks \citep{wang2025affordbot,li2026a3r,zhao2026cov}. GREP applies this principle to scene-specific inspection: each check gathers the measurements and views needed for judgment, and each verdict retains its supporting evidence. Adaptive evidence collection implements the inspection guided by our evaluation philosophy.

\section{GREP Evaluation Framework}
\label{sec:method}

GREP evaluates generated rooms from their assets, positions, orientations, scales, and available boundaries. We first outline the \emph{evaluation workflow}, from scene analysis to evidence collection and reporting. We then describe the \emph{evaluation dimensions}, which make the three principles operational through eight metrics, the \emph{evidence operations} that support adaptive inspection, and the resulting \emph{scores and reports}.

\subsection{GREP Workflow}
\label{sec:sieve-workflow}

\begin{figure}[t]
\centering
\includegraphics[width=\linewidth]{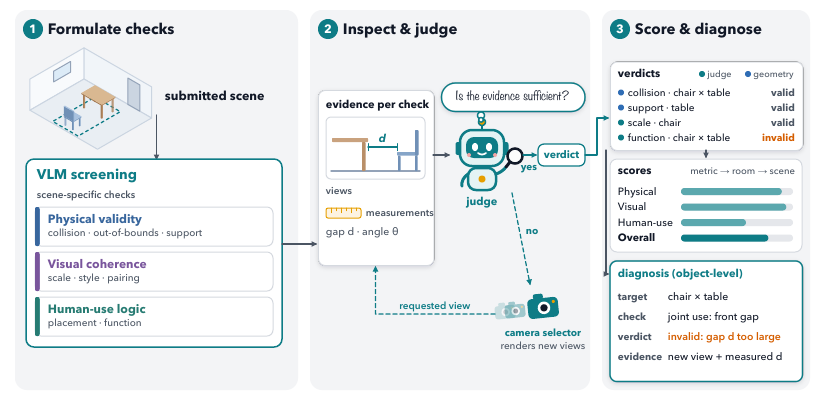}
\vspace{-15pt}
\caption{
GREP evaluation framework. (1) Analyze the room globally to raise questions under the three principles. (2) Adaptively gather evidence to resolve remaining checks. (3) Aggregate scores and compile a diagnostic report.}
\vspace{-15pt}
\label{fig:framework}
\end{figure}

Figure~\ref{fig:framework} summarizes the evaluation in three stages: formulating checks, inspecting and judging them, and scoring with diagnoses; the evaluation procedure depends on the metric (Appendix~\ref{app:workflow}).
(1) \textit{Step 1: Analyze the room to raise questions.}
\label{sec:sieve-obligations}
The evaluation agent uses a vision-language model (VLM) to inspect room-wide views, object information, and intended use. It identifies objects and relationships requiring inspection under the three principles and formulates scene-specific \textit{checks}. Screening resolves conclusive cases at this stage: geometric predicates settle clear physical cases, and VLM screens clear evidently valid Scale and Pairing cases (Appendix~\ref{app:workflow}).
(2) \textit{Step 2: Complete the remaining checks.}
For each unresolved check, the agent gathers relevant evidence for judgment. It records a verdict with its reasons and supporting evidence.
(3) \textit{Step 3: Aggregate scores and compile the report.}
GREP aggregates the judgments into metric, room, and scene scores (Section~\ref{sec:sieve-scoring}). The accompanying report records affected objects, reasons, evidence, and unresolved gaps, providing specific guidance for scene revision.

\subsection{Evaluation Dimensions}
\label{sec:sieve-metrics}

The three principles provide conceptual guidance for evaluation. Eight metrics make these principles operational. Each metric specifies a quality dimension; each check applies it to particular objects or relationships, and each check type specifies the property to inspect.

\noindent\textbf{Checks for the physical validity.}
A room must first obey physical constraints. \emph{Collision} checks for unintended intersections, distinguishing them from intended attachments. \emph{Out-of-bounds} checks whether objects cross the supplied spatial boundary. \emph{Support} checks whether objects have physical support. Together, these checks assess basic physical validity; they do not determine whether the arrangement looks coherent or supports its intended use.

\noindent\textbf{Checks for the visual coherence.}
Physically valid objects can still form a visually inconsistent room. \emph{Scale} checks whether sizes are reasonable for the object categories, \emph{Style} checks visual compatibility, and \emph{Pairing} checks whether object choices are compatible or unnecessarily repetitive. These checks assess how the selected assets fit together in size, style, and composition.

\noindent\textbf{Checks for the human-use logic.}
A visually coherent room must also make sense for human use. \emph{Placement} asks whether people would place an object in that location, considering its properties and scene context. \emph{Function} then asks whether objects can be used as intended in that arrangement, checking access, orientation, and spacing for individual objects and objects used together.

\subsection{Evidence Operations}
\label{sec:sieve-evidence}

Guided by the check types above, the evaluation agent formulates scene-specific questions and gathers evidence to answer them. We allow three types of operations that the agent selects and combines as needed.
(1) \textit{Measure spatial relations.}
Compute distances, relative angles, overlap, and clearance for the target objects from their geometry and poses.
(2) \textit{Retrieve existing views.}
Each check starts from the room-wide views and the views already acquired for its scope; Function checks in the same object group share one evidence bank. This tool returns views of the target objects and their surroundings without additional rendering.
(3) \textit{Render additional views.}
When existing views do not reveal a required feature, acquire another view by changing the camera while keeping the scene fixed. The evaluator checks whether the image supplies the missing evidence, and the view is kept for later checks in the same scope.

These tools support all three principles. If geometry cannot settle a physical check, for example, the agent can inspect a view of the relevant region or request a new one if the region is hidden. Visual coherence and human-use checks likewise obtain evidence as needed. Evidence collection ends when it is sufficient or the acquisition budget is exhausted; unresolved gaps remain in the report. Appendix~\ref{app:workflow} details measurement procedures, camera selection, and budget controls.

\subsection{Scores and Reports}
\label{sec:sieve-scoring}

\noindent\textbf{Scores.}
GREP computes scores from confirmed defects identified by its checks.
For each metric, repeated observations of the same defect are
deduplicated into defect events. Each event receives a severity-based
penalty, which is attributed to the responsible objects.
The metric score combines the total object penalty $B_m$,
normalized by the room's object count $N$, with the largest
event penalty $P_m$:
\[
  S_m = 100(1-\alpha P_m)(1-B_m/N)^\gamma,
  \qquad \alpha=0.4,\quad \gamma=4.
\]
Higher scores indicate fewer or less severe defects.
Appendix~\ref{app:scoring} specifies the event penalties,
object attribution, and aggregation rules.

Metric scores are combined using predefined nominal weights
to obtain the room score. We also report principle scores as
weight-normalized averages of the metrics belonging to physical
validity, visual coherence, and human-use logic.
The \frozenassets{} configuration excludes visual coherence
metrics from the overall score and renormalizes the remaining
weights. For multi-room scenes, room scores are aggregated
using net floor area as the weight. Model and method results
then average the included scene scores equally.

\noindent\textbf{Reports.}
The report pairs these score summaries with the underlying inspection records.
Each record corresponds to a check, identifying the check type, target objects or relations, verdict, reason, and supporting measurements or views. Evidence gaps and decision sources thus remain explicit. This links the overall assessment to specific issues and provides the generator with context for scene revision, as we will show by experiments.

\section{Experiments}
\label{sec:experiments}
We first use GREP and specially designed benchmarks to evaluate current scene generation methods, identify performance trends, and derive design insights. We then focus on GREP itself, examining its value for scene refinement and its reliability as an evaluator.

\subsection{Benchmark Observations}

\subsubsection{Benchmark Protocol}
GREP uses GPT-5.6 Sol as the evaluation agent. We organize baselines into three generation settings and describe their protocols and task suites below.
We report scores for each principle and the overall scene score, abbreviating physical validity, visual coherence, and human-use logic as Physical, Visual, and Human-use, respectively.
Further details appear in Appendices~\ref{app:benchmark} and~\ref{app:generation-protocols}.

(1) \textbf{Model-native planning.}
This setting focuses on native reasoning and planning abilities of language models. Given an open space or a floorplan with walls, a model first describes the objects it intends to include. An external matcher retrieves assets matching these descriptions from the Imaginarium asset library \citep{zhu2025imaginarium}. The model then outputs their scales, positions, and orientations in a single call. We evaluate GPT-6 Astra, GPT-5.6 Sol, Claude Opus 5, Claude Fable 5, Claude Sonnet 5, Kimi K3, GLM-5.3, and HY4 Preview. The suite contains 30 scenes: 20 open spaces and 10 floorplans with relatively simple rectangular wall layouts splitting the space into multiple rooms.

(2) \textbf{Specialized pipelines.}
This setting evaluates specialized pipelines that use handcrafted rules to guide and adjust model-generated asset arrangements. We compare LayoutGPT \citep{layoutgpt2023}, DirectLayout \citep{ran2025direct}, LayoutVLM \citep{layoutvlm2024}, and SceneWeaver \citep{yang2025sceneweaver} against direct placement (\emph{None}), using both the stronger GPT-5.6 Sol and the weaker GPT-4.1 as backbones.
The suite contains ten open-space scenes. Following common practice, we use \frozenassets{}, providing each scene with a fixed set of preselected assets that every method must use and place. Visual coherence checks (Scale, Style, and Pairing) are excluded from Overall because asset choices and scales are predetermined.

(3) \textbf{Harnessed agents.}
This setting evaluates language models within a shared agent harness. We use GPT-5.6 Sol, Kimi K3, GLM-5.3, and HY4 Preview as backbones. Models have access to information about the full Imaginarium database (2,043 3D assets across 385 categories) and callable tools: \texttt{get-task}, \texttt{search-assets}, \texttt{inspect-asset}, \texttt{validate-submission}, and \texttt{finalize-submission}. The harness supports tool use, self-correction, and iterative refinement without exposing GREP scores.
However, the harness also imposes \textit{strict constraints}. Its suite contains ten scenes with irregular wall boundaries, predefined room uses, and prescribed asset counts, which may restrict the model's planning flexibility.
For evaluating these advanced systems, we use ten highly complex floorplans with irregular wall layouts.

\begin{table}[t]
  \centering
  \small
  \setlength{\tabcolsep}{2pt}
  \caption{Model-native results in open spaces (without walls) and floorplans with rectangular wall layouts. Bold marks the highest value in each row. HY4 Preview's with-wall average covers nine of the ten floorplans (Appendix~\ref{app:eval-config}).}
  \vspace{5pt}
  \label{tab:model}
  \begin{tabular*}{\textwidth}{@{\extracolsep{\fill}}lcccccccc@{}}
    \toprule
    & \shortstack{GPT-6\\Astra}
    & \shortstack{GPT-5.6\\Sol}
    & \shortstack{Claude\\Opus 5}
    & \shortstack{Claude\\Fable 5}
    & \shortstack{Claude\\Sonnet 5}
    & \shortstack{Kimi\\K3}
    & \shortstack{GLM-\\5.3}
    & \shortstack{HY4\\Preview} \\
    \hline\hline
    \multicolumn{9}{@{}l}{\emph{Open space}} \\
    Overall    & \textbf{84.5} & 73.7 & 81.5 & 77.9 & 65.3 & 73.2 & 70.8 & 61.8 \\
    Physical  & 85.0 & 81.0 & \textbf{90.3} & 77.8 & 72.4 & 70.2 & 67.2 & 55.9 \\
    Visual    & 88.4 & 84.8 & \textbf{95.3} & 95.1 & 87.0 & 91.2 & 90.2 & 86.8 \\
    Human-use  & \textbf{83.4} & 66.4 & 72.4 & 74.8 & 56.0 & 72.0 & 69.7 & 61.4 \\
    \addlinespace
    \multicolumn{9}{@{}l}{\emph{With wall layouts}} \\
    Overall    & \textbf{81.2} & 73.4 & 69.5 & 72.5 & 65.7 & 66.6 & 68.7 & 54.2 \\
    Physical  & \textbf{80.2} & 73.8 & 79.4 & 79.2 & 73.5 & 75.4 & 72.0 & 54.0 \\
    Visual    & 86.9 & 89.9 & 94.0 & 91.5 & \textbf{96.7} & 92.9 & 89.5 & 87.5 \\
    Human-use  & \textbf{80.9} & 69.9 & 57.6 & 64.0 & 54.0 & 55.1 & 62.4 & 47.9 \\
    \bottomrule
  \end{tabular*}
\end{table}

\begin{table}[t]
  \centering
  \small
  \setlength{\tabcolsep}{4pt}
  \caption{Results of specialized pipelines across model backbones and asset placement methods. Visual coherence metrics are excluded under \frozenassets{}. ``None'' means raising the asset arrangement plans with the native planning abilities of models without requiring a specialized pipeline. Bold marks the highest value in each row.}
  \vspace{5pt}
  \label{tab:pipeline}
  \begin{tabular}{@{}lccccc@{}}
    \toprule
    & None & LayoutGPT & LayoutVLM & DirectLayout & SceneWeaver \\
    \hline\hline
    \multicolumn{6}{@{}l}{\emph{GPT-5.6 Sol backbone}} \\
    Overall    & \textbf{78.3} & 74.3 & 73.6 & 68.5 & 62.1 \\
    Physical  & \textbf{83.4} & 81.1 & 81.3 & 81.3 & 77.5 \\
    Human-use  & \textbf{74.5} & 69.4 & 68.0 & 59.2 & 50.9 \\
    \addlinespace
    \multicolumn{6}{@{}l}{\emph{GPT-4.1 backbone}} \\
    Overall    & 40.2 & 31.0 & \textbf{54.5} & 44.1 & 40.0 \\
    Physical  & 36.6 & 19.0 & \textbf{57.9} & 38.1 & 29.9 \\
    Human-use  & 42.8 & 39.7 & \textbf{52.1} & 48.5 & 47.4 \\
    \bottomrule
  \end{tabular}
\end{table}

\begin{table}[t]
  \centering
  \small
  \setlength{\tabcolsep}{5pt}
  \caption{Results of harnessed agents under default requirements and a stricter condition requiring 20\% more assets. Bold marks the highest value in each row.}
  \label{tab:agent}
  \vspace{5pt}
  \begin{tabular}{@{}lcccc@{}}
    \toprule
    & GPT-5.6 Sol & Kimi K3 & GLM-5.3 & HY4 Preview \\
    \hline\hline
    \multicolumn{5}{@{}l}{\emph{Default constraints}} \\
    Overall    & 83.0 & \textbf{88.5} & 84.7 & \textbf{88.5} \\
    Physical  & 75.8 & 78.5 & 84.5 & \textbf{89.6} \\
    Visual    & \textbf{98.8} & 97.3 & 98.0 & 97.8 \\
    Human-use  & 85.2 & \textbf{94.0} & 82.3 & 86.0 \\
    \addlinespace
    \multicolumn{5}{@{}l}{\emph{20\% more assets}} \\
    Overall    & 80.3 & 87.4 & 88.3 & \textbf{89.3} \\
    Physical  & 66.8 & 73.2 & 84.3 & \textbf{89.4} \\
    Visual    & \textbf{98.8} & 98.3 & 96.7 & 96.0 \\
    Human-use  & 86.6 & \textbf{95.7} & 89.7 & 87.9 \\
    \bottomrule
  \end{tabular}
\end{table}

\subsubsection{Benchmark Results}
\label{sec:results}
\label{sec:findings}
GREP supports three observations on key questions in scene generation: the impact of native ability of models, the effectiveness of specialized pipelines, and the value of agent harnesses.

\textbf{(1) Model-native planning}

\begin{figure}[t]
  \centering
  \includegraphics[width=\linewidth]{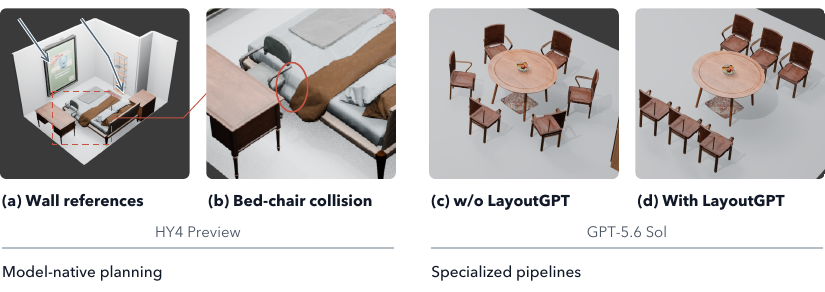}
  \caption{Qualitative examples of Model-native planning and pipeline-dependent
  generation. (a) HY4 Preview uses walls as placement references,
  (b) yet its layout contains a bed-chair collision, shown in the magnified
  view. With the same dining assets and GPT-5.6 Sol
  backbone, (c) Model-native arranges chairs around the round table,
  while (d) LayoutGPT places them in two straight rows.}
  \label{fig:observations}
\vspace{-10pt}
\end{figure}

\textit{Results.} Stronger models, like GPT-5.6 Sol, already produce largely sensible layouts with relatively few conflicts, achieving an overall score above 70 in both settings (Table~\ref{tab:model}). Weaker models such as HY4 Preview remain less capable despite using spatial cues from walls. Generation traces show HY4 using wall positions and inward normals to place and orient the picture and bookshelf (Figure~\ref{fig:observations}(a)), yet its layout still contains a bed-chair collision (Figure~\ref{fig:observations}(b)).

\textit{Insights.} With few constraints, recent advances in planning abilities already enable stronger models to generate sensible room layouts.

\textbf{(2) Specialized pipelines}

\textit{Results.} However, strong planning does not ensure that a model can recognize unsuitable guidance. Handcrafted pipelines can improve weaker models' plans: LayoutVLM raises GPT-4.1's Overall from 40.2 to 54.5 (Table~\ref{tab:pipeline}). For GPT-5.6 Sol, such guidance may instead disrupt an otherwise sensible plan. Figure~\ref{fig:observations}(c--d) shows Sol arranging chairs around a round table on its own, but placing them in two straight rows under LayoutGPT. All four pipelines lower Sol's Overall by 4.0--16.2 points, suggesting that the model does not adequately assess whether their guidance suits the scene.

\noindent\textit{Insights.} Better planning alone is not enough. Models still \textit{cannot effectively assess the suitability of external guidance}, which undermines the coherence of the overall scene.

\textbf{(3) Harnessed agents}

\begin{figure}[t]
  \centering
  \includegraphics[width=\linewidth]{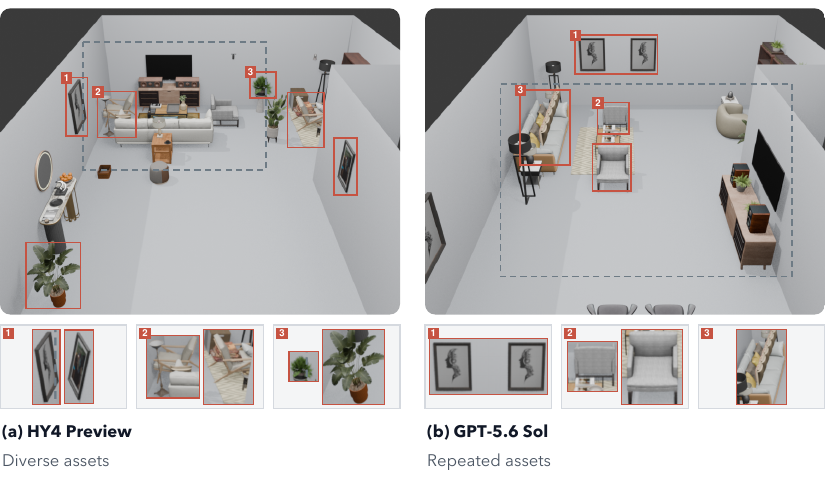}
\vspace{-15pt}

  \caption{Qualitative examples under the agent harness.
(a) HY4 Preview uses a more diverse set of assets.
(b) GPT-5.6 Sol repeats several assets, including wall pictures,
chairs, and sofa cushions. Numbered boxes mark the enlarged insets below
each layout: (1) wall pictures, (2) chairs, and (3) plants in (a) or sofa
cushions in (b).}
\vspace{-15pt}
  \label{fig:agent}
\end{figure}

\textit{Results.} The agent setting reveals a striking ranking reversal: HY4 Preview outperforms GPT-5.6 Sol, reversing their order in the Model-native setting (Tables~\ref{tab:model} and~\ref{tab:agent}). One possible explanation is that harness compensates for weaker planning, while requirements such as mandatory asset counts interfere with stronger models' own plans. To probe sensitivity to these requirements, we increase the required asset count by 20\% within the agent suite. HY4's Overall improves slightly, whereas Sol's deficit relative to HY4 widens from 5.5 to 9.0 points and its Physical falls from 75.8 to 66.8, revealing the strong harness hurts the strong model performances.
Figure~\ref{fig:agent} illustrates differences in asset selection: HY4 Preview uses a more diverse set of assets, whereas GPT-5.6 Sol uses more repeated assets.

\noindent\textit{Insights.} Stronger planning now comes with \textit{poorer adaptation to external constraints}: models may adhere to existing plans rather than effectively revise them to meet new requirements.

\textbf{Summary}

These observations suggest a progression from \textit{reliance}, through \textit{conflict}, toward \textit{integration}. Weaker models can benefit from external guidance that compensates for limited planning. Stronger models can form better plans of their own, yet struggle to reconcile them with supplied guidance and constraints. The next stage should unite these strengths: models should judge when guidance is useful, adapt it to the scene, and incorporate constraints into a coherent overall plan. This calls for training and system design that develop planning ability and flexibility together.

\subsection{Evaluating GREP}
\label{sec:validation}

We examine three aspects of GREP itself: the value of its diagnostic reports, its agreement with human judgments, and its repeatability. Further details and analyses are provided in the appendix.

\subsubsection{Diagnostic Feedback for Revision}
\label{sec:findings-feedback}

We show that GREP's diagnostic reports provide more useful guidance for scene revision than scores alone.
Specifically, we test their value on five scenes using GPT-4.1. Each branch revises the generated scene three times. At each revision, the model receives the current scene and GREP's scene score or report as context. We have the following observations.

\noindent\textit{(1) Reports improve revision beyond scores.}
After three revisions, the Report branch raises the mean score
from 0.585 to 0.676, while score-only and no-feedback baselines
show slight declines (Figure~\ref{fig:feedback}(a)).
This study uses the refinement-specific scoring protocol
specified in Appendix~\ref{app:feedback-protocol}.

\noindent\textit{(2) Reports prompt concrete changes.}
Across the three revision rounds and five scenes, the diagnostic
branch changes the layout in all 15 revisions, compared with
5/15 for scores alone and 6/15 without feedback. These results suggest that numerical scores provide little direction for revision, whereas diagnoses identify specific issues that the model can act on.

The example in Figure~\ref{fig:feedback}(b) shows a dining layout
before and after report-guided revision. Several chairs overlap
in the initial layout; the revised layout separates them
around the table.

\begin{figure}[t]
  \centering
\includegraphics[width=\linewidth]{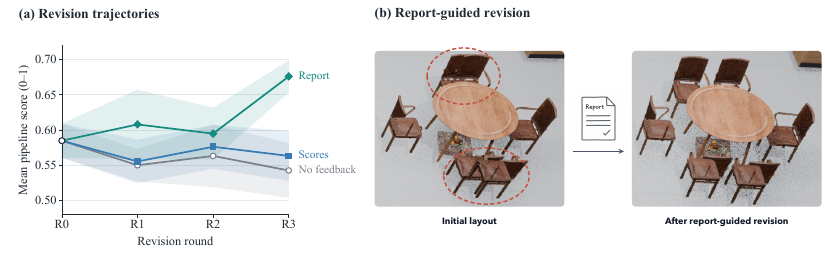}
\vspace{-20pt}
  \caption{Report-guided scene refinement with GPT-4.1.
  (a) Mean score across five scenes from the shared initial
  layout (R0) through three revisions (R1--R3), using the
  0--1 pipeline scoring protocol in
  Appendix~\ref{app:feedback-protocol}; shaded bands show $\pm 1$
  standard error across the five scenes.
  (b) The dining layout of S101 before and after report-guided revision:
  dashed circles mark overlapping chairs, which are separated in the
  revised layout.
  These views were rendered again from the saved layouts.}
  \label{fig:feedback}
\vspace{-10pt}

\end{figure}

\begin{figure}[t]
  \centering
  \includegraphics[width=\linewidth]{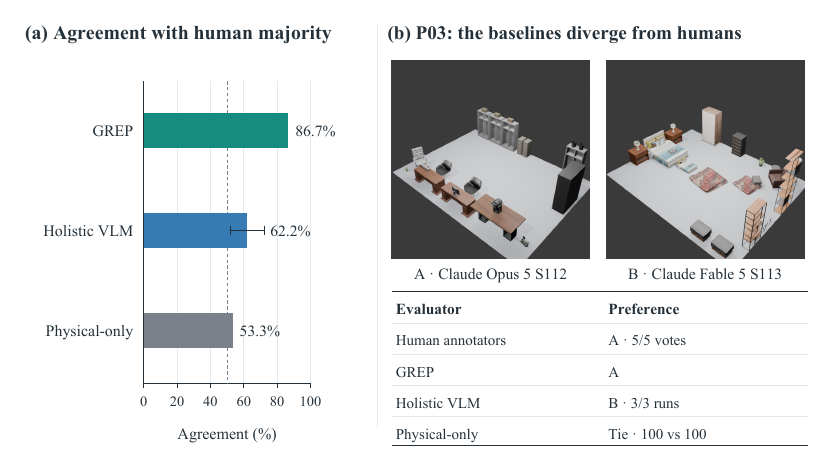}
\caption{Agreement with human overall preferences across 15 pairs
of six scenes. (a) Agreement with the human majority; the error bar
shows the between-run standard deviation of the Holistic VLM, and the
dashed line marks chance (50\%).
(b) Pair P03: all five annotators and GREP prefer A,
all three Holistic VLM runs prefer B, and Physical-only
assigns equal scores.}
  \label{fig:evaluators}
\vspace{-10pt}
\end{figure}

\subsubsection{Agreement with Human Judgments}
Figure~\ref{fig:evaluators} compares GREP, Holistic VLM judgments,
and Physical-only scores with human overall preferences.
GREP agrees with the human majority on 86.7\% of pairs,
compared with 62.2\% for Holistic VLM averaged over three runs
and 53.3\% for Physical-only.
Here, we reuse GREP's Physical score as the Physical-only baseline, as it already covers typical physical checks.
Appendices~\ref{app:human} and~\ref{app:baselines}
provide the study procedure and baseline details.

\subsubsection{Repeatability}
We evaluated 12 selected open-space scenes five times each with fixed inputs, configuration, and judge. The per-scene standard deviation of the overall score has a median of 2.9 points and a maximum of 7.3. Differences between scenes are much larger: repeat-to-repeat variation accounts for 8\% of the score variance across these scenes (Appendix~\ref{app:repeatability}). A smaller-scale analysis with a second judge, Claude Fable 5, on six of these scenes preserves the ordering of the scenes (Spearman $\rho=0.94$) but scores all six higher, by 4.8 points on average (paired $t$-test, $p=0.12$, $n=6$).

\section{Limitations}
\label{sec:limitations}
\textit{Single judge model.} All reported evaluations use GPT-5.6 Sol as the judge, and GPT-5.6 Sol is also one of the evaluated generators. LLM evaluators can recognize and favor their own generations \citep{panickssery2024selfpreference}. Our judge-sensitivity analysis with Claude Fable 5 covers only six scenes and cannot establish whether either judge favors its own model family (Appendix~\ref{app:repeatability}).

\textit{Coverage of discovered checks.} Discovery instantiates checks from fixed check types over the submitted objects. Judgment coverage refers to these instantiated checks; it does not show that discovery found every relevant object or relationship in a room (Appendix~\ref{app:contracts}).

\textit{Forced verdicts under a bounded budget.} Evidence acquisition is bounded. Checks still unresolved when the budget is exhausted receive a forced binary verdict from the retained evidence, and on some scene-quality procedures non-hard failures resolve to a labeled default-valid outcome (Appendix~\ref{app:adjudication}). Each record identifies these decision sources, but such verdicts rest on incomplete evidence and can be wrong.

\section{Conclusion}
\label{sec:conclusion}
GREP grounds generated room evaluation in three progressive principles: physical validity, visual coherence, and human-use logic. These principles guide autonomous inspection and adaptive evidence collection, yielding multidimensional scores and diagnostic reports.  GREP achieves 87\%
agreement with human majority preferences. Under the refinement-specific
scoring protocol, report feedback raises the mean score from 0.585 to
0.676 after three revisions, while score-only and no-feedback baselines
show slight declines. Our benchmark observations suggest that stronger planning alone does not ensure effective use of guidance or constraints. This requires models and systems that combine strong autonomous planning with the flexibility to incorporate guidance and constraints into coherent scene plans.

\subsection*{AI Use Statement}
We used AI tools to assist with writing, literature search, research ideation, and execution. The authors reviewed and verified all content, analyses, and claims.

\subsection*{Ethics Statement}
We identify no significant ethical risks specific to this work beyond those generally associated with generative AI and automated evaluation.

\subsection*{Reproducibility Statement}
The paper and appendix provide the evaluation procedure, experimental settings, and implementation details needed to reproduce our results.

\bibliography{references}

\begin{thebibliography}{29}
\providecommand{\natexlab}[1]{#1}
\providecommand{\url}[1]{\texttt{#1}}
\expandafter\ifx\csname urlstyle\endcsname\relax
  \providecommand{\doi}[1]{doi: #1}\else
  \providecommand{\doi}{doi: \begingroup \urlstyle{rm}\Url}\fi

\bibitem[{Blender Online Community}(2026)]{blender}
{Blender Online Community}.
\newblock Blender -- a {3D} modelling and rendering package.
\newblock Blender Foundation, \url{https://www.blender.org}, 2026.

\bibitem[Cho et~al.(2024)Cho, Hu, Baldridge, Garg, Anderson, Krishna, Bansal, Pont-Tuset, and Wang]{cho2024dsg}
Jaemin Cho, Yushi Hu, Jason Baldridge, Roopal Garg, Peter Anderson, Ranjay Krishna, Mohit Bansal, Jordi Pont-Tuset, and Su~Wang.
\newblock Davidsonian scene graph: Improving reliability in fine-grained evaluation for text-to-image generation.
\newblock In \emph{The Twelfth International Conference on Learning Representations}, 2024.
\newblock URL \url{https://openreview.net/forum?id=ITq4ZRUT4a}.

\bibitem[Feng et~al.(2023)Feng, Zhu, Fu, Jampani, Akula, He, Basu, Wang, and Wang]{layoutgpt2023}
Weixi Feng, Wanrong Zhu, Tsu-Jui Fu, Varun Jampani, Arjun Akula, Xuehai He, Sugato Basu, Xin~Eric Wang, and William~Yang Wang.
\newblock {LayoutGPT}: Compositional visual planning and generation with large language models.
\newblock In \emph{Advances in Neural Information Processing Systems}, volume~36, 2023.
\newblock URL \url{https://arxiv.org/abs/2305.15393}.

\bibitem[Hessel et~al.(2021)Hessel, Holtzman, Forbes, Bras, and Choi]{hessel2021clipscore}
Jack Hessel, Ari Holtzman, Maxwell Forbes, Ronan~Le Bras, and Yejin Choi.
\newblock {CLIPScore}: A reference-free evaluation metric for image captioning.
\newblock In \emph{Proceedings of the 2021 Conference on Empirical Methods in Natural Language Processing}, pp.\  7514--7528, 2021.
\newblock \doi{10.18653/v1/2021.emnlp-main.595}.

\bibitem[Hu et~al.(2023)Hu, Liu, Kasai, Wang, Ostendorf, Krishna, and Smith]{hu2023tifa}
Yushi Hu, Benlin Liu, Jungo Kasai, Yizhong Wang, Mari Ostendorf, Ranjay Krishna, and Noah~A. Smith.
\newblock {TIFA}: Accurate and interpretable text-to-image faithfulness evaluation with question answering.
\newblock In \emph{Proceedings of the IEEE/CVF International Conference on Computer Vision}, pp.\  20406--20417, 2023.

\bibitem[Kang et~al.(2026)Kang, Choi, Hwangbo, Lim, Ong, and Yeo]{legoeval2026}
Minseok Kang, Dongwook Choi, Gyeom Hwangbo, Seungwon Lim, {Kai Tzu-iunn} Ong, and Jinyoung Yeo.
\newblock Apples on the table? {Evaluating} text-guided {3D} scene synthesis via fine-grained constraint verification.
\newblock In \emph{Proceedings of the 2026 Conference on Empirical Methods in Natural Language Processing}, 2026.
\newblock URL \url{https://arxiv.org/abs/2511.03001}.

\bibitem[Lee et~al.(2024)Lee, Kim, Park, Kim, and Seo]{lee2024prometheusvision}
Seongyun Lee, Seungone Kim, Sue Park, Geewook Kim, and Minjoon Seo.
\newblock {Prometheus-Vision}: Vision-language model as a judge for fine-grained evaluation.
\newblock In \emph{Findings of the Association for Computational Linguistics: ACL 2024}, pp.\  11286--11315, 2024.
\newblock \doi{10.18653/v1/2024.findings-acl.672}.

\bibitem[Li et~al.(2026)Li, Feng, Li, Shang, Zheng, Dong, and Shi]{li2026a3r}
Di~Li, Jie Feng, Guanbin Li, Ronghua Shang, Yuhui Zheng, Weisheng Dong, and Guangming Shi.
\newblock {A3R}: Agentic affordance reasoning via cross-dimensional evidence in {3D} {Gaussian} scenes.
\newblock \emph{arXiv preprint arXiv:2604.01882}, 2026.

\bibitem[Lin \& Mu(2024)Lin and Mu]{lin2024instructscene}
Chenguo Lin and Yadong Mu.
\newblock {InstructScene}: Instruction-driven {3D} indoor scene synthesis with semantic graph prior.
\newblock In \emph{The Twelfth International Conference on Learning Representations}, 2024.

\bibitem[Maillard et~al.(2026)Maillard, Engelmann, Durand, Pan, You, Litany, Guibas, and Ovsjanikov]{maillard2026sceneteract}
L{\'e}opold Maillard, Francis Engelmann, Tom Durand, Boxiao Pan, Yang You, Or~Litany, Leonidas Guibas, and Maks Ovsjanikov.
\newblock {SceneTeract}: Agentic functional affordances and {VLM} grounding in {3D} scenes.
\newblock \emph{arXiv preprint arXiv:2603.29798}, 2026.

\bibitem[Mao et~al.(2025)Mao, Zhong, Fang, Zheng, Tang, Zhu, Tan, and Zhou]{mao2025spatiallm}
Yongsen Mao, Junhao Zhong, Chuan Fang, Jia Zheng, Rui Tang, Hao Zhu, Ping Tan, and Zihan Zhou.
\newblock {SpatialLM}: Training large language models for structured indoor modeling.
\newblock In \emph{Advances in Neural Information Processing Systems}, volume~38, 2025.

\bibitem[Panickssery et~al.(2024)Panickssery, Bowman, and Feng]{panickssery2024selfpreference}
Arjun Panickssery, Samuel~R. Bowman, and Shi Feng.
\newblock {LLM} evaluators recognize and favor their own generations.
\newblock In \emph{Advances in Neural Information Processing Systems}, volume~37, 2024.

\bibitem[Paschalidou et~al.(2021)Paschalidou, Kar, Shugrina, Kreis, Geiger, and Fidler]{paschalidou2021atiss}
Despoina Paschalidou, Amlan Kar, Maria Shugrina, Karsten Kreis, Andreas Geiger, and Sanja Fidler.
\newblock {ATISS}: Autoregressive transformers for indoor scene synthesis.
\newblock In \emph{Advances in Neural Information Processing Systems}, volume~34, 2021.

\bibitem[Ran et~al.(2025)Ran, Li, Xu, Yu, and Dai]{ran2025direct}
Xingjian Ran, Yixuan Li, Linning Xu, Mulin Yu, and Bo~Dai.
\newblock Direct numerical layout generation for {3D} indoor scene synthesis via spatial reasoning.
\newblock In \emph{Advances in Neural Information Processing Systems}, volume~38, 2025.
\newblock URL \url{https://arxiv.org/abs/2506.05341}.

\bibitem[Sun et~al.(2025)Sun, Liu, Gu, Lim, Bhat, Tombari, Li, Haber, and Wu]{layoutvlm2024}
Fan-Yun Sun, Weiyu Liu, Siyi Gu, Dylan Lim, Goutam Bhat, Federico Tombari, Manling Li, Nick Haber, and Jiajun Wu.
\newblock {LayoutVLM}: Differentiable optimization of {3D} layout via vision-language models.
\newblock In \emph{Proceedings of the IEEE/CVF Conference on Computer Vision and Pattern Recognition}, pp.\  29469--29478, 2025.

\bibitem[Tam et~al.(2026)Tam, Pun, Wang, Chang, and Savva]{tam2026sceneeval}
Hou In~Ivan Tam, Hou In~Derek Pun, Austin~T. Wang, Angel~X. Chang, and Manolis Savva.
\newblock {SceneEval}: Evaluating semantic coherence in text-conditioned {3D} indoor scene synthesis.
\newblock In \emph{Proceedings of the IEEE/CVF Winter Conference on Applications of Computer Vision}, pp.\  7355--7365, 2026.

\bibitem[Tang et~al.(2024)Tang, Nie, Markhasin, Dai, Thies, and Nie{\ss}ner]{tang2024diffuscene}
Jiapeng Tang, Yinyu Nie, Lev Markhasin, Angela Dai, Justus Thies, and Matthias Nie{\ss}ner.
\newblock {DiffuScene}: Denoising diffusion models for generative indoor scene synthesis.
\newblock In \emph{Proceedings of the IEEE/CVF Conference on Computer Vision and Pattern Recognition}, 2024.

\bibitem[Wan et~al.(2026)Wan, Mao, Wu, Xie, Xiang, Wang, Zhang, Liu, Dai, and Liu]{wan2026nala}
Cheng Wan, Yongsen Mao, Wenzheng Wu, Yuxuan Xie, Chucheng Xiang, Runze Wang, Xiang Zhang, Zhongyuan Liu, Rushi Dai, and Yuan Liu.
\newblock {NaLA}: A {3D} native {LLM} layout agent for high-quality {3D} scene generation.
\newblock In \emph{European Conference on Computer Vision}, 2026.
\newblock URL \url{https://arxiv.org/abs/2606.29395}.

\bibitem[Wang et~al.(2025)Wang, Yang, Xu, Wu, Li, and Zhao]{wang2025affordbot}
Xinyi Wang, Xun Yang, Yanlong Xu, Yuchen Wu, Zhen Li, and Na~Zhao.
\newblock {AffordBot}: {3D} fine-grained embodied reasoning via multimodal large language models.
\newblock In \emph{Advances in Neural Information Processing Systems}, volume~38, 2025.
\newblock \doi{10.52202/085713-4691}.

\bibitem[Xiang et~al.(2026)Xiang, Bao, Feng, Wu, Liu, Guan, and Liu]{xiang2026co}
Chucheng Xiang, Ruchao Bao, Biyin Feng, Wenzheng Wu, Zhongyuan Liu, Yirui Guan, and Ligang Liu.
\newblock Co-layout: Llm-driven co-optimization for interior layout.
\newblock In \emph{Proceedings of the AAAI Conference on Artificial Intelligence}, volume~40, pp.\  14371--14379, 2026.

\bibitem[Yang et~al.(2024{\natexlab{a}})Yang, Jia, Zhi, and Huang]{physcene2024}
Yandan Yang, Baoxiong Jia, Peiyuan Zhi, and Siyuan Huang.
\newblock {PhyScene}: Physically interactable {3D} scene synthesis for embodied {AI}.
\newblock In \emph{Proceedings of the IEEE/CVF Conference on Computer Vision and Pattern Recognition}, pp.\  16262--16272, 2024{\natexlab{a}}.

\bibitem[Yang et~al.(2025)Yang, Jia, Zhang, and Huang]{yang2025sceneweaver}
Yandan Yang, Baoxiong Jia, Shujie Zhang, and Siyuan Huang.
\newblock {SceneWeaver}: All-in-one {3D} scene synthesis with an extensible and self-reflective agent.
\newblock In \emph{Advances in Neural Information Processing Systems}, volume~38, 2025.
\newblock URL \url{https://arxiv.org/abs/2509.20414}.

\bibitem[Yang et~al.(2024{\natexlab{b}})Yang, Sun, Weihs, VanderBilt, Herrasti, Han, Wu, Haber, Krishna, Liu, Callison-Burch, Yatskar, Kembhavi, and Clark]{holodeck2024}
Yue Yang, Fan-Yun Sun, Luca Weihs, Eli VanderBilt, Alvaro Herrasti, Winson Han, Jiajun Wu, Nick Haber, Ranjay Krishna, Lingjie Liu, Chris Callison-Burch, Mark Yatskar, Aniruddha Kembhavi, and Christopher Clark.
\newblock {Holodeck}: Language guided generation of {3D} embodied {AI} environments.
\newblock In \emph{Proceedings of the IEEE/CVF Conference on Computer Vision and Pattern Recognition}, pp.\  16227--16237, 2024{\natexlab{b}}.

\bibitem[Zechner \& {Earendil Works}(2026)Zechner and {Earendil Works}]{zechner2026pi}
Mario Zechner and {Earendil Works}.
\newblock {Pi Coding Agent}.
\newblock \url{https://github.com/earendil-works/pi}, 2026.
\newblock npm package \texttt{@earendil-works/pi-coding-agent}.

\bibitem[Zhang et~al.(2026)Zhang, Jones, Wu, and Agrawala]{zhang2026weakverifiers}
Sharon Zhang, R.~Kenny Jones, Jiajun Wu, and Maneesh Agrawala.
\newblock Aggregating {LLM}-based weak verifiers for spatial layout generation.
\newblock \emph{arXiv preprint arXiv:2606.05268}, 2026.
\newblock \doi{10.48550/arXiv.2606.05268}.
\newblock URL \url{https://arxiv.org/abs/2606.05268}.

\bibitem[Zhang et~al.(2025)Zhang, Li, Long, Zhang, Lin, Yang, Xie, Yang, Liu, Lin, Huang, and Zhou]{zhang2025qwen3embedding}
Yanzhao Zhang, Mingxin Li, Dingkun Long, Xin Zhang, Huan Lin, Baosong Yang, Pengjun Xie, An~Yang, Dayiheng Liu, Junyang Lin, Fei Huang, and Jingren Zhou.
\newblock {Qwen3 Embedding}: Advancing text embedding and reranking through foundation models.
\newblock \emph{arXiv preprint arXiv:2506.05176}, 2025.

\bibitem[Zhao et~al.(2026)Zhao, Liu, Zhang, Wang, Chen, Zhu, Haffari, and Zhuang]{zhao2026cov}
Haoyu Zhao, Akide Liu, Zeyu Zhang, Weijie Wang, Feng Chen, Ruihan Zhu, Gholamreza Haffari, and Bohan Zhuang.
\newblock {CoV}: Chain-of-view prompting for spatial reasoning.
\newblock In \emph{Findings of the Association for Computational Linguistics: ACL 2026}, pp.\  32427--32440, 2026.
\newblock \doi{10.18653/v1/2026.findings-acl.1623}.

\bibitem[Zheng et~al.(2023)Zheng, Chiang, Sheng, Zhuang, Wu, Zhuang, Lin, Li, Li, Xing, Zhang, Gonzalez, and Stoica]{zheng2023judging}
Lianmin Zheng, Wei-Lin Chiang, Ying Sheng, Siyuan Zhuang, Zhanghao Wu, Yonghao Zhuang, Zi~Lin, Zhuohan Li, Dacheng Li, Eric~P. Xing, Hao Zhang, Joseph~E. Gonzalez, and Ion Stoica.
\newblock Judging {LLM}-as-a-judge with {MT}-bench and chatbot arena.
\newblock In \emph{Advances in Neural Information Processing Systems}, volume~36, 2023.

\bibitem[Zhu et~al.(2025)Zhu, Huang, Xie, Deng, Yu, Guan, Liu, Zhu, Zhao, Liu, and Zeng]{zhu2025imaginarium}
Xiaoming Zhu, Xu~Huang, Qinghongbing Xie, Zhi Deng, Junsheng Yu, Yirui Guan, Zhongyuan Liu, Lin Zhu, Qijun Zhao, Ligang Liu, and Long Zeng.
\newblock {Imaginarium}: Vision-guided high-quality {3D} scene layout generation.
\newblock \emph{arXiv preprint arXiv:2510.15564}, 2025.

\end{thebibliography}
\bibliographystyle{venue/iclr2027/iclr2027_conference}

\appendix

\section{Evaluation Details}
\label{app:workflow}

This appendix expands the three stages of Section~\ref{sec:method}: formulating checks, inspecting and judging them, and producing scores with diagnoses. A \emph{metric} is one of the eight evaluation dimensions under the physical, visual coherence, and human-use principles. A \emph{check} is a scene-specific question about particular objects or relationships. A \emph{check type} specifies the property to inspect, and an \emph{evaluation procedure} specifies how a metric formulates and resolves its checks. The agent's work is divided among VLM discovery, deterministic procedures, a camera selector, and a metric evaluator that issues the verdicts for each metric. Appendix~\ref{app:scoring} gives the score computation, and Appendix~\ref{app:provenance} lists the evaluation configuration.

\subsection{End-to-End Evaluation Overview}
\label{app:end-to-end}

\begin{figure}[h]
  \centering
  \includegraphics[width=\linewidth]{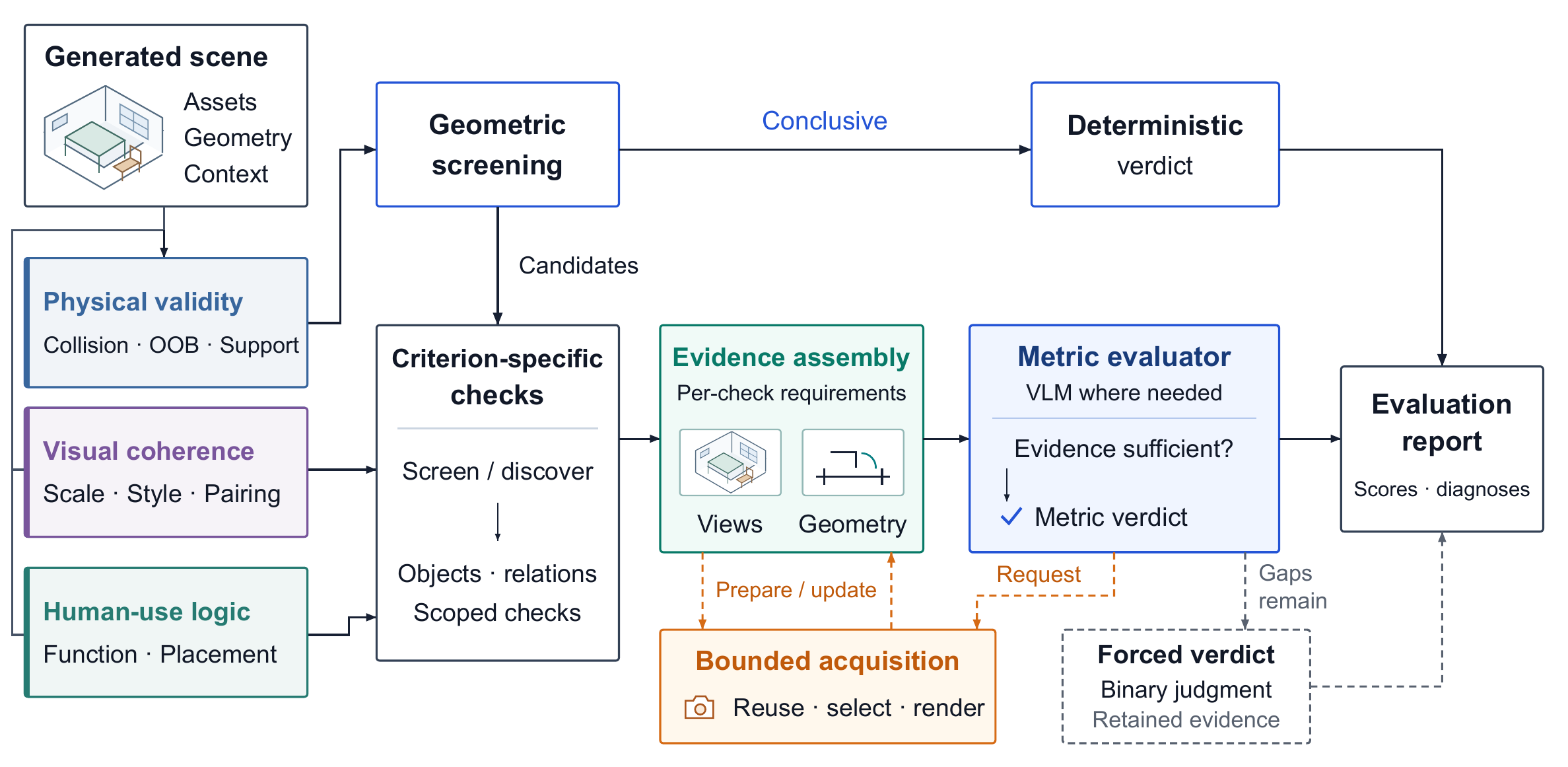}
  \caption{Per-metric evaluation procedures. Criterion-specific checks organize
  evidence for adjudication; conclusive geometric predicates are resolved
  directly. Bounded acquisition supports check preparation and requests for
  more evidence. The procedure depends on the metric (Table~\ref{tab:app-metric-routes}). OOB denotes out-of-bounds.}
  \label{fig:workflow-detail}
\end{figure}

GREP evaluates each room from its objects, asset geometry, positions, orientations, scales, and available scene context. Relationships and explanations authored by the generator are treated as claims, not as evidence that the described relations were realized. Before evaluation, input checks verify that the scene is parseable, that coordinates and units are consistent, and that the required asset data are available. A room that fails these checks is not evaluated. Boundary checks use the spatial boundary defined for the evaluated setting.

Table~\ref{tab:app-metric-routes} summarizes how each metric formulates, scopes, and resolves its checks, and Fig.~\ref{fig:workflow-detail} shows the corresponding procedures. Physical metrics begin with geometric screening, which resolves conclusive cases directly and passes the remaining candidates on for inspection. Scale and Pairing begin with a VLM screen of structured object information that resolves clearly valid cases and sends suspected defects to visual confirmation. Style begins with a global visual screen, followed by local confirmation when the global screen identifies a suspected defect. Function and Placement use VLM discovery to select, from the submitted objects, the objects and relationships to inspect within fixed check types.

Checks that need further inspection receive measurements and views. The metric evaluator interprets this evidence and may request additional views when a required property remains unclear, and the camera selector supplies them while the scene stays fixed. Verdicts come from deterministic procedures or from the metric evaluator, never from the camera selector, and a check may be resolved without passing through every component. Each check ends in an inspection record that retains its verdict, reason, supporting evidence, and decision source. These records feed both score aggregation and object-level diagnoses.

\begin{table}[h]
  \centering
  \small
  \caption{Per-metric evaluation procedures. A group is an object group from the room
  partition (Appendix~\ref{app:instantiation}), and the evaluator is the metric's
  VLM evaluator.}
  \label{tab:app-metric-routes}
  \newcommand{\appcell}{\raggedright\hspace{0pt}}%
  \begin{tabular}{@{}lp{3.5cm}p{3.1cm}p{3.6cm}@{}}
    \toprule
    Metric & Check formulation & Evidence scope & Resolution \tabularnewline
    \midrule
    Collision & \appcell Mesh and oriented bounding box (OBB) screening of object pairs & \appcell Object pair & \appcell Deterministic for certified separation or a bounded interface overlap, else evaluator \tabularnewline
    Out-of-bounds & \appcell Screening against the room envelope & \appcell Object and boundary & \appcell Deterministic for containment within tolerance, else evaluator \tabularnewline
    Support & \appcell Support-chain screening along gravity & \appcell Object and its supports & \appcell Deterministic for grounded contact, else evaluator \tabularnewline
    \addlinespace
    Scale & \appcell VLM screen of structured object information & \appcell Group, after the screen & \appcell Screen for clearly valid cases, else evaluator \tabularnewline
    Style & \appcell Global visual screen & \appcell Room, then group & \appcell Screen, or evaluator after local confirmation \tabularnewline
    Pairing & \appcell VLM screen of the object inventory & \appcell Room inventory, at most one global view & \appcell Screen for clearly valid cases, else evaluator \tabularnewline
    \addlinespace
    Function & \appcell Two-stage VLM discovery of affordances and joint-use relations & \appcell Group with a shared evidence bank, plus own scopes for cross-group relations & \appcell Evaluator \tabularnewline
    Placement & \appcell VLM discovery of location checks & \appcell Group, with the global view & \appcell Evaluator \tabularnewline
    \bottomrule
  \end{tabular}
\end{table}

\subsection{Check Instantiation and Evidence Scope}
\label{app:instantiation}

A metric names the quality to assess, while a check names the objects, relationship, and property to inspect in a particular room. In the chair--table example of Fig.~\ref{fig:framework}, Function checks whether the chair is oriented toward the table and whether the space between them permits their joint use.

Function discovery runs in two VLM stages over the global view, object identities and categories, and supplied scene and architectural context. The first stage records the affordances of individual objects, and the second audits joint-use relationships between them. Discovery selects among the submitted objects and does not add objects. Accepted proposals take one of five check types: architecture orientation, clearance, local affordance confirmation, directional correspondence, and relative-use geometry. Placement discovery proposes checks of three types (support surface and height, scene zone, and contextual anchor) from the global view, object information, and the spatial context produced by Function check discovery. This context directs attention and does not carry Function verdicts.

Deterministic procedures validate the object references and check type of each proposal, remove duplicates, and convert accepted proposals into checks. Each check fixes the inspection target and its evidence requirements and leaves the verdict to the metric evaluator. Discovery therefore instantiates checks from a fixed vocabulary rather than generating new evaluation procedures.

Object groups define local inspection scopes. A VLM proposes the partition of the room, with a geometric-topology fallback when the proposal fails. The partition considers support and attachment, local interaction, and shared anchors before proximity alone, keeps unusual objects, and gives functional relationships that span groups their own scopes. Table~\ref{tab:app-metric-routes} lists the scope each metric uses, and all scopes stay within the evaluated room.

\subsection{Check-Specific Evidence Requirements}
\label{app:contracts}

Each check specifies the observations and measurements its judgment needs. For Function and Placement, each check type carries predefined requirements, supplemented by observation goals from discovery and by later requests from the metric evaluator. A chair--table orientation check requires identifiable targets, a reference for the chair's usable side, and the relative orientation and distance between the two objects; a clearance check additionally requires evidence of the surrounding region. The usable-side reference comes from the asset catalog where available and otherwise from VLM interpretation of asset-side previews. Combined with the object transforms, it gives the world-space directions used for measurement.

An evidence packet may contain object identities, asset geometry, global and local views, spatial measurements, and scene context, and it records unavailable measurements and unmet observation requirements. Function checks in the same object group draw on a shared evidence bank, so a view acquired for one check can serve the others in that group.

Evidence availability and sufficiency are assessed separately. A deterministic gate confirms only that an observation can be consumed: the file exists, decodes, and is not blank. Whether the observation reveals the orientation, contact region, or obstruction relevant to the check is judged by the metric evaluator. Evidence coverage refers to the instantiated checks and does not show that discovery found every relevant relationship in the room.

\subsection{Bounded Evidence Acquisition}
\label{app:acquisition}

Evidence is acquired at two points: while Function checks are prepared, and when a metric evaluator requests more evidence during adjudication. For Function, deterministic measurements are attached to accepted checks before any camera scheduling, so the spatial facts remain available even when no view is added. These measurements use the asset geometry and the resolved usable-side reference. Clearance is measured as the obstruction of an approach corridor in front of the usable side.

Function check preparation first reserves views for relationships that span object groups and then allocates the remaining probe units by the additional observation coverage each view provides. The resulting views enter the group's shared evidence bank. The other metrics acquire views for their own scopes (Table~\ref{tab:app-metric-routes}).

Camera candidates are generated deterministically and filtered for geometric feasibility, visibility, and framing. The camera selector returns views rather than verdicts, and the metric evaluator decides whether a new view resolves the question.

Acquisition limits apply per evaluation episode to evidence rounds, images, camera-selection calls, and camera actions, and they count usage carried into the episode. Function check preparation has its own budget, so no single rendering budget applies to a whole room. Budget exhaustion and infeasible views are recorded in the inspection record. Table~\ref{tab:app-acquisition-limits} in Appendix~\ref{app:provenance} lists the limits.

\subsection{Adjudication and Reporting}
\label{app:adjudication}

Conclusive geometric predicates are resolved deterministically. They include certified separation between two objects, containment within the evaluated boundary up to a floor-contact tolerance, grounded support contact, and two bounded interface overlaps: a millimeter-scale overlap at a support interface, and a vertical-only overlap between an explicit floor covering and an object resting on it. An overlap between coarse bounding volumes identifies a collision candidate rather than a verdict, and relationship claims authored by the generator never exempt a pair.

When a physical check remains inconclusive, the metric evaluator adjudicates it from scoped geometric and visual evidence and treats the measurements as fixed facts. For human-use checks, measurements establish spatial facts that the evaluator interprets in context: a chair--table distance informs the judgment, but usability also depends on orientation and access. Checks still unresolved when the evidence budget is exhausted receive a forced binary verdict from the retained evidence, and on some scene-quality procedures non-hard failures resolve to a labeled default-valid outcome.

Each inspection record identifies the affected objects or relationship, the check, the verdict and its reason, the supporting views or measurements, and the decision source, which distinguishes deterministic verdicts, evaluator verdicts, forced verdicts, and policy defaults. Execution failures are recorded separately and are not treated as verdicts. Appendix~\ref{app:scoring} specifies how the records contribute to scores, and the same records provide the object-level diagnoses used for scene revision.

\section{Scoring Details}
\label{app:scoring}

We compute quality scores in three stages: assigning penalties to
confirmed defects, combining these penalties for each object and metric,
and aggregating the resulting metric scores.

Let $m$ denote an evaluation metric, $\mathcal{O}$ the set of objects in
the evaluated room, and $N = |\mathcal{O}|$ the number of objects, with
$N > 0$. Each object instance is counted once, and all metrics use the
same $N$.

A confirmed defect constitutes an event $e$ with penalty $p(e)$.
We define $p_0$ as the minimum penalty and $p_1$ as the maximum penalty
for a confirmed defect. The penalty range is $\Delta p = p_1 - p_0$.
We use $p_0 = 0.4$ and $p_1 = 1.0$, giving $\Delta p = 0.6$.

For Collision, Support, and Out-of-bounds, let $M(e) \in [0,1]$
denote the normalized geometric severity of event $e$. The event
penalty is
\begin{equation}
    p(e) = p_0 + \Delta p\,M(e).
    \label{eq:app-event-penalty}
\end{equation}
$M(e)$ reflects the extent of penetration, the support gap, or the
extent of boundary crossing. When a defect is confirmed but its
magnitude cannot be reliably measured, we assign $p_0$ and record
the missing measurement. The remaining metrics use the severity
levels in Table~\ref{tab:app-severity-penalties}.

\begin{table}[htbp]
    \centering
    \small
    \caption{Severity levels and their event penalties.}
    \label{tab:app-severity-penalties}
    \begin{tabular}{@{}ll@{}}
        \toprule
        Metric & Severity and penalty \\
        \midrule
        Scale     & Noticeable: $p_0$; gross: $p_1$ \\
        Style     & Noticeable: $p_0$; gross: $p_1$ \\
        Pairing   & Invalid: $p_1$ \\
        Function  & Impaired: $p_0$; blocked: $p_1$ \\
        Placement & Atypical: $p_0$; implausible: $p_1$ \\
        \bottomrule
    \end{tabular}
\end{table}

Valid judgments contribute no penalty. Repeated observations of the
same defect are merged, retaining the strongest confirmed severity.
Let $\mathcal{E}_m$ denote the resulting set of defect events for
metric $m$.

Each event penalty is assigned to the responsible objects.
Let $a(e,o)$ denote the portion of event $e$'s penalty assigned to
object $o$. An event involving one scoring object assigns its entire
penalty to that object; a shared penalty is divided among the scoring
objects. The allocation preserves the event's total penalty:
\begin{equation}
    \sum_{o \in \mathcal{O}} a(e,o) = p(e),
    \qquad a(e,o) \geq 0.
    \label{eq:app-penalty-allocation}
\end{equation}
Objects receiving no part of the event's penalty have $a(e,o)=0$.
For example, a Collision event divides its penalty equally between
its two endpoints.

We define $b(m,o)$ as the combined penalty assigned to object $o$
under metric $m$. For Collision, Support, Out-of-bounds, Scale,
Style, and Pairing, assigned penalties are added and capped at one:
\begin{equation}
    b(m,o)
    =
    \min\left(
        1,\,
        \sum_{e \in \mathcal{E}_m} a(e,o)
    \right).
    \label{eq:app-object-penalty-sum}
\end{equation}
For Function and Placement, we retain the largest assigned penalty:
\begin{equation}
    b(m,o)
    =
    \max_{e \in \mathcal{E}_m} a(e,o).
    \label{eq:app-object-penalty-max}
\end{equation}
Thus, two impaired Function defects assigned entirely to the same
object contribute $p_0$ in total. If either defect is blocked, the
object's contribution becomes $p_1$. An object with no assigned
defects has $b(m,o)=0$.

We then compute two quantities for each metric. The total object
penalty $B_m$ measures the accumulated defects across the room:
\begin{equation}
    B_m = \sum_{o \in \mathcal{O}} b(m,o).
    \label{eq:app-total-penalty}
\end{equation}
The maximum event penalty $P_m$ measures the severity of the single
worst defect before its penalty is divided among objects:
\begin{equation}
    P_m = \max_{e \in \mathcal{E}_m} p(e).
    \label{eq:app-maximum-penalty}
\end{equation}
For example, an event with penalty one contributes $P_m=1$ even
when its penalty is divided between two objects. When
$\mathcal{E}_m$ is empty, we define all corresponding object
penalties and both $B_m$ and $P_m$ to be zero.

Two hyperparameters control the score mapping:
$\alpha \in [0,1]$ controls the influence of the worst defect,
and $\gamma > 0$ controls sensitivity to the overall penalty
proportion $B_m/N$. We use $\alpha=0.40$ and $\gamma=4$.
The metric score combines these two quantities:
\begin{equation}
    S_m
    =
    100
    \left(1-\alpha P_m\right)
    \left(1-\frac{B_m}{N}\right)^{\gamma}.
    \label{eq:app-metric-score}
\end{equation}
Scores lie between zero and 100, with a fully evaluated metric
receiving 100 when no defects are found. Placement uses its
primary checks for this calculation; the additional global review
does not contribute.

For example, a room containing 20 objects with one impaired
Function defect assigned to one object has $B_m=p_0$ and $P_m=p_0$.
Its Function score is
\begin{equation}
    S_{\mathrm{Function}}
    =
    100
    \left(1-\alpha p_0\right)
    \left(1-\frac{p_0}{20}\right)^{\gamma}
    \approx 77.48.
    \label{eq:app-score-example}
\end{equation}

The room score is a weighted average of its metric scores.
Let $w_m$ denote the weight of metric $m$, with
$\sum_m w_m=1$:
\begin{equation}
    Q_{\mathrm{room}} = \sum_m w_m S_m.
    \label{eq:app-room-score}
\end{equation}
The metric weights are listed in Table~\ref{tab:app-metric-weights}.

\begin{table}[htbp]
    \centering
    \small
    \caption{Metric weights used to aggregate room scores.}
    \label{tab:app-metric-weights}
    \begin{tabular}{@{}lr@{}}
        \toprule
        Metric & $w_m$ \\
        \midrule
        Collision     & 0.15 \\
        Support       & 0.15 \\
        Out-of-bounds & 0.08 \\
        Scale         & 0.03 \\
        Style         & 0.03 \\
        Pairing       & 0.04 \\
        Function      & 0.32 \\
        Placement     & 0.20 \\
        \bottomrule
    \end{tabular}
\end{table}

The result tables also report one score per principle, the weight-normalized average of its metric scores:
\begin{equation}
  S_p = \frac{\sum_{m \in p} w_m S_m}{\sum_{m \in p} w_m}.
  \label{eq:app-principle-score}
\end{equation}
The physical principle comprises Collision, Support, and Out-of-bounds; the visual coherence principle comprises Scale, Style, and Pairing; and the human-use principle comprises Function and Placement. The room score therefore equals $0.38\,S_{\mathrm{phys}} + 0.10\,S_{\mathrm{vis}} + 0.52\,S_{\mathrm{use}}$. The \frozenassets{} configuration fixes asset identities and scales before placement. Since object selection, style, and scale are predetermined in these experiments, their overall score excludes the visual coherence metrics. We mark these experiments with an asterisk ($*$), giving an overall score of $(0.38\,S_{\mathrm{phys}} + 0.52\,S_{\mathrm{use}})/0.90$.
Multi-room scene scores are weighted averages of room scores
using net floor area. Model and method scores are equal averages
across included scenes.

Unresolved judgments and evaluation failures are recorded separately
from valid and invalid judgments. Partial results retain their judgment-coverage information and enter the aggregate only when they meet the experiment's scoring eligibility criteria. Missing results
are never assigned zero or a perfect score.

\section{Experimental Setup}
\label{app:provenance}

\subsection{Benchmark Composition}
\label{app:benchmark}

The benchmark comprises three generation settings, with task counts
specified separately for each setting.

\paragraph{Model-native (Table~\ref{tab:model}).}
Each model is evaluated on 30 tasks: 20 open-space scenes
(S100--S119) and 10 relatively simple floorplans
(\texttt{layout\_01}--\texttt{layout\_10}). The open-space setting
(WF, wall-free) has one room per scene and no enclosing
walls. The floorplan tasks form the rectangular (RC) setting,
comprising 31 rectangular rooms across the ten floorplans.

\paragraph{Specialized pipelines (Table~\ref{tab:pipeline}).}
Each model--pipeline configuration is evaluated on ten open-space
scenes (S100--S109), comprising ten single rooms under
\frozenassets{}. These tasks share their room briefs and boundaries
with the corresponding WF subset of the Model-native setting, while their
asset inventories are fixed before generation.

\paragraph{Harnessed agents (Table~\ref{tab:agent}).}
Each backbone is evaluated on 20 cases: ten under the default
condition and ten under the \emph{more complicated condition}
($+20\%$ assets). The two conditions reuse the same ten irregular
floorplans, with 42 rooms in each condition. We refer to this
architectural setting as irregular (IR). The increased furnishing
requirements define the second set of cases; the floorplans and
required room programs are shared between the two conditions.

WF task descriptions specify the intended room use, dimensions, target instance-count range, and furnishing instruction. RC additionally provides room rectangles, local-to-global offsets, active walls, openings, and the order in which rooms are generated. Each RC room has a prescribed use and object-count range. Counts refer to expanded object instances rather than object categories or rows in an object plan.

The IR floorplans are adapted from the SpatialLM dataset \citep{mao2025spatiallm}. Their horizontal coordinates are uniformly scaled by a factor of 1.25, preserving the polygon structure while increasing floor area by a factor of 1.5625. The benchmark supplies the required room programs and activities, together with total and per-room instance-count ranges. It does not supply the source dataset's room-to-use assignment: the generator assigns each required program to a room. The architecture, program requirements, count ranges, and asset-scale constraints remain fixed within a comparison.

The shared Imaginarium asset library \citep{zhu2025imaginarium} contains 2,043 distinct assets and 385 distinct labels in its \texttt{category} field. Assets are accompanied by descriptions, dimensions, and identity metadata. \frozenassets{} fixes a human-curated inventory for each of ten WF tasks (S100--S109), comprising 269 object instances drawn from 168 distinct assets in total. Every pipeline receives the same inventory for a task, including exact asset identities and native scales. This condition isolates placement from object and asset selection.

\subsection{Generation Settings}
\label{app:generation-protocols}

\paragraph{Model-native.}
We use \emph{Plan--Retrieve--Place} for ordinary Model-native generation. It uses two model calls: one to plan the objects and one to place them, with deterministic asset retrieval between the calls. The retrieval profile uses Qwen3-Embedding-0.6B \citep{zhang2025qwen3embedding} and a deterministic Top-1 policy combining semantic similarity with size compatibility and a fixed tie-breaking order. The model has no interactive tool loop or evaluator feedback during these two generation stages. Task descriptions, retrieval policy, and output contracts are held fixed across the backbones compared under this generation setting.

RC is generated sequentially, one room at a time. Each room receives one object-planning call and one placement call, with retrieval between them. Assembly prefixes instance identifiers with the room identity and translates room-local coordinates into the floorplan frame; it does not reposition, rotate, rescale, or repair the objects.

GPT-6 Astra, GPT-5.6 Sol, and HY4 Preview use \texttt{high} reasoning effort. Claude Opus 5, Sonnet 5, and Fable 5 use the provider-default \texttt{high} setting. Kimi K3 and GLM-5.3 use \texttt{max}, with a lower-effort fallback when reasoning exceeds the token limit. The Model-native setting supplies the model-controlled planning and placement reference.

\paragraph{Specialized pipelines.}
We pair LayoutGPT, DirectLayout, LayoutVLM, and SceneWeaver with GPT-5.6 Sol and GPT-4.1 under \frozenassets{}. Each method receives one WF task, its room boundary, fixed object inventory, asset identities, and physical dimensions. Adapters preserve these bindings while translating between the benchmark scene representation and each method's native inputs and outputs.

LayoutGPT uses its numerical Cascading Style Sheets (CSS) representation with fixed object dimensions and an in-context example set. DirectLayout receives the frozen assets through its native generation and refinement pipeline. LayoutVLM receives fixed asset sizes and identities and runs its constraint-generation and layout-solving procedure. SceneWeaver initializes the supplied meshes and retains its iterative reflection loop, while object insertion, removal, replacement, and resizing are disabled. These adaptations constrain the inventory while preserving the methods' placement procedures.

The within-backbone baseline, labeled None in Table~\ref{tab:pipeline}, uses the placement stage of Plan--Retrieve--Place. Because the inventory is already fixed, it makes one placement call and skips object planning and retrieval, with no iterative refinement. Methods retain their own call and optimization schedules, so inference calls, token use, and wall-clock computation are not equalized. Since asset selection and scale are fixed, the overall \frozenassets{} score excludes the visual coherence metrics and follows the renormalization in Appendix~\ref{app:scoring}.

\paragraph{Harnessed agents.}
All backbones operate through a common harness built on the open-source Pi coding agent \citep{zechner2026pi} (version 0.85.0), with an isolated workspace, immutable architecture, required room programs, count ranges, and access to the same asset database. The agent chooses its object composition, searches for assets, inspects their metadata, and writes a complete submission. The tool interface provides \texttt{get-task}, \texttt{search-assets}, \texttt{inspect-asset}, \texttt{validate-submission}, and \texttt{finalize-submission}. Asset inspection returns structured metadata and bounding-box dimensions. The harness provides no rendering or image observations to the generation agent.

The public validator checks the submission schema, layout identity, room-program assignment, total and per-room instance counts, asset membership, binding and placement cardinality, and unit asset scale. It supplies neither GREP quality scores nor defect diagnoses or suggested repairs. A completed generation requires a finalized submission that passes this contract. The same scene definitions, asset access, and episode limits apply across the agent backbones.

Each episode permits at most 120 model requests and 160 benchmark-tool calls, with sublimits of 80 asset searches, 80 asset inspections, and 30 submission validations. A search returns at most 12 candidates. The episode wall-clock limit is 14,400 seconds. These are generation limits, separate from the evaluation budgets in Section~\ref{app:eval-config}.

The default furnishing condition uses the benchmark's prescribed integer ranges. The \emph{more complicated condition} increases the required furnishing density: each per-room lower and upper bound is multiplied by 1.2 and rounded down, and scene-level bounds are the sums of the resulting room bounds. The architecture, room programs, asset scales, and tool limits remain unchanged. The default condition provides the reference for harnessed agents, while the more complicated condition tests increased furnishing demands.

\subsection{Evaluation Configuration}
\label{app:eval-config}

\paragraph{Evaluation and rendering.}
GREP evaluates generated scenes room by room, using GPT-5.6 Sol for VLM judgments. Evidence rendering uses the supplied asset meshes and scene poses, with the scene fixed throughout inspection. The reference rendering configuration uses the EEVEE rendering engine of Blender \citep{blender} with $768\times768$ final views and $256\times256$ candidate previews. Global room views initialize inspection, and targeted views provide additional evidence when requested.  The metric procedures are described in Appendix~\ref{app:workflow}, and Table~\ref{tab:app-acquisition-limits} gives the evidence-acquisition limits. Score parameters, metric weights, \frozenassets{} renormalization, and room-to-scene aggregation follow Appendix~\ref{app:scoring}.

\begin{table}[h]
  \centering
  \small
  \caption{Evidence-acquisition limits. Episode limits apply to each scoped
  evaluation episode, and Function check preparation has a separate budget.}
  \label{tab:app-acquisition-limits}
  \begin{tabular}{@{}lr@{}}
    \toprule
    Limit & Value \\
    \midrule
    Evidence rounds per episode & 3 \\
    Images per episode & 8 \\
    Views per round & 2 \\
    Camera-selection calls per episode & 4 \\
    Camera actions per episode & 3 \\
    Deterministic camera candidates per request & 6 \\
    VLM repair proposals per request & 3 \\
    Function check preparation: probe units & 32 \\
    Function check preparation: candidates per probe & 4 \\
    Function check preparation: views kept per probe & 1 \\
    Function shared evidence bank: images per active window & 6 \\
    \bottomrule
  \end{tabular}
\end{table}

\paragraph{Completion and inclusion.}
Generation completion, report completion, and judgment coverage are tracked separately. A failed generation has no quality score, including when it exhausts its allowed attempts without producing an accepted scene. Missing or excluded results are not assigned zero or a perfect score. Only scenes with a scoring-eligible report for every required room enter the aggregate. Room scores are weighted by their net floor areas within each scene, and eligible scenes receive equal weight.

For Table~\ref{tab:model}, all eight models contribute 20 scoring-eligible
WF scenes. In RC, seven models contribute all ten floorplans
(31 rooms); HY4 Preview contributes nine floorplans (27 rooms).
Its generated \texttt{layout\_09} is excluded because one required
room lacks a scoring-eligible report. For Table~\ref{tab:pipeline},
all ten scenes are generated and included for each model--pipeline
configuration. For Table~\ref{tab:agent}, every backbone contributes
all ten IR floorplans (42 rooms) under both conditions, and all 168
rooms in each condition have scoring-eligible reports.

\section{Evaluator Validation Details}
\label{app:validation}

This appendix gives the experimental procedures and results behind
Section~\ref{sec:validation}: repeatability under a fixed configuration
and sensitivity to the judge model (Appendix~\ref{app:repeatability}),
agreement with human judgments (Appendix~\ref{app:human}), and a
comparison with alternative evaluators (Appendix~\ref{app:baselines}).
GREP scene scores use the 0--100 scale defined in
Appendix~\ref{app:scoring}; agreement rates below are percentages
of pairwise preferences.

\subsection{Repeatability and Judge Sensitivity}
\label{app:repeatability}

\paragraph{Experimental procedure.}
We analyze 12 selected scenes generated by eight Model-native backbones on the open-space tasks S110--S119. Each scene was evaluated five times with GPT-5.6 Sol as the judge, with the generation inputs, evaluator configuration, and judge fixed. The inputs were re-hashed before every repeat and were identical across repeats, and all 60 evaluations completed with full judgment coverage, so the differences reported below arise from evaluation alone.

\paragraph{Scene scores.}
Table~\ref{tab:app-repeatability} lists the mean, standard deviation (SD), and range of each scene over the five repeats. The per-scene standard deviation has a median of 2.9 points (from 0.4 to 7.3), and the median difference between the highest and lowest repeat is 7.2 points (maximum 17.2). Differences between scenes are larger: repeat-to-repeat variation accounts for 8\% of the score variance across the 12 scenes (intraclass correlation 0.92), and any two repeats order the scenes with a Spearman correlation between 0.70 and 0.98 (median 0.90). In every scene, at least one metric's number of confirmed defects changed across repeats.

\begin{table}[h]
  \centering
  \small
  \caption{Repeatability of GREP scores. Each scene was evaluated five times with fixed inputs, configuration, and judge (GPT-5.6 Sol).}
  \label{tab:app-repeatability}
  \begin{tabular}{@{}llrrr@{}}
    \toprule
    Generator & Task & Mean & SD & Min--Max \\
    \midrule
    GPT-6 Astra & S112 & 97.7 & 2.0 & 94.4--99.3 \\
    Kimi K3 & S118 & 95.8 & 5.2 & 86.6--98.7 \\
    GPT-5.6 Sol & S118 & 94.4 & 0.4 & 93.9--94.8 \\
    Claude Fable 5 & S110 & 92.1 & 0.7 & 90.9--92.4 \\
    Claude Opus 5 & S114 & 89.1 & 2.8 & 84.1--90.7 \\
    Claude Sonnet 5 & S116 & 88.1 & 1.0 & 87.1--89.6 \\
    GPT-6 Astra & S113 & 87.6 & 7.1 & 74.9--91.1 \\
    GLM-5.3 & S118 & 83.5 & 2.2 & 81.0--85.7 \\
    HY4 Preview & S113 & 80.8 & 6.0 & 73.3--88.3 \\
    GPT-5.6 Sol & S116 & 64.4 & 4.6 & 58.5--70.9 \\
    Claude Opus 5 & S111 & 63.4 & 7.3 & 54.5--71.7 \\
    Claude Fable 5 & S114 & 54.9 & 2.9 & 51.1--58.9 \\
    \bottomrule
  \end{tabular}
\end{table}

\paragraph{Implications for comparisons.}
Assuming independent evaluations, the variation observed here gives a standard error of about 1.3 points for a mean over ten scenes and about 1.9 points for the difference between two such means. Differences between model means below about 4 points are therefore within what a single evaluation per scene can produce. The refinement study reports paired differences with their standard errors for the same reason (Appendix~\ref{app:feedback}).

\paragraph{Judge sensitivity.}
We use 26 room evaluations across six scenes to assess judge sensitivity, with Claude Fable 5 as the alternative judge. The two judges order the six scenes almost identically (Spearman $\rho=0.94$; only the two lowest-scoring scenes swap). Fable 5 scores every scene higher, by 4.8 points on average (paired $t=1.88$ with five degrees of freedom, $p=0.12$). The largest gap, 16.8 points, occurs on a scene generated by Claude Fable 5; the other five differ by 0.3 to 6.7 points. Its repeated scores also vary less (median per-scene standard deviation 0.1 versus 2.9 points for GPT-5.6 Sol on the same scenes). With six scenes, the analysis indicates that a second judge preserves relative comparisons while shifting absolute scores; it cannot establish whether a judge favors scenes generated by its own model family.

\subsection{Human Agreement}
\label{app:human}

\paragraph{Experimental procedure.}
Five computer graphics researchers, including graduate students and professors, independently compared six selected open-space scenes. The study contains all 15 unordered scene pairs, with each scene appearing in five pairs. Each pair presents one global perspective image per scene, its intended room use and dimensions, and descriptions of physical validity, visual coherence, and human-use logic. Annotators selected A or B for the scene with better overall quality in its intended room context. They did not explore the scenes or receive local views, metric scores, or per-check judgments. All five annotators completed all 15 pairs, yielding 75 preferences.

\paragraph{Results.}
All five annotators agreed on 11 of the 15 pairs; the other four pairs split three to two. Across the ten annotator pairs for each scene pair, agreement was 126/150, or 84.0\% (Fleiss' $\kappa=0.677$). GREP's Overall scene scores agreed with the human majority on 13/15 pairs (86.7\%) and with 61/75 individual preferences (81.3\%). These comparisons use the same six scenes, whose 15 pairwise decisions are not independent scene samples.

\paragraph{Disagreements.}
GREP differed from the human majority on two pairs. For P07, all five annotators preferred GLM-5.3 S115 over Claude Opus 5 S111, whereas GREP's Overall scores preferred the latter. For P09, three annotators preferred GLM-5.3 S115 and two preferred Claude Fable 5 S113, whereas GREP preferred Fable S113. Thus, P07 is a unanimous human disagreement and P09 a divided human judgment.

\subsection{Comparison with Alternative Evaluators}
\label{app:baselines}

Fig.~\ref{fig:evaluators} compares GREP with a holistic VLM and a Physical-only score baseline using the same 15 pairs. The holistic VLM uses GPT-5.6 Sol with the same model configuration as GREP's judge. In each of three separate runs, it receives the same six global images, room context, and overall-preference instructions as the human annotators, and returns one A/B choice for every pair. It does not receive scene geometry, GREP scores, additional views, or other annotators' choices. Across the three runs, its agreement with the human majority was 62.2\% $\pm$ 10.2\%, and agreement with individual preferences was 60.4\% $\pm$ 12.7\% (mean $\pm$ between-run SD). Its choice was identical across runs on 10/15 pairs.

The Physical-only baseline orders each pair by its existing GREP Physical score, which combines Collision, Support, and Out-of-bounds. It agreed with the human majority on 8/15 pairs (53.3\%) and with 36/75 individual preferences (48.0\%). Its scores tied on P03; because human annotators made a forced A/B choice, a tied prediction receives no agreement credit. These are post-adjudication GREP Physical scores, not the output of an independent geometry-only detector. Humans and the holistic VLM used the same global-image evidence bank, whereas GREP and the Physical-only baseline reuse the original GREP evaluations and their evidence.

\section{Refinement Study Details\texorpdfstring{\textsuperscript{*}}{*}}
\label{app:feedback}

This appendix documents the refinement study of Section~\ref{sec:findings-feedback}: the experimental procedure and feedback content (Appendix~\ref{app:feedback-protocol}), per-scene results (Appendix~\ref{app:feedback-results}), layout changes (Appendix~\ref{app:feedback-changes}), an example of diagnosed defects and their revision (Appendix~\ref{app:feedback-case}), and the evaluation records (Appendix~\ref{app:feedback-records}).

\subsection{Experimental Procedure and Feedback Content}
\label{app:feedback-protocol}

The study uses five open-space scenes (S101--S105) under \frozenassets{}, with GPT-4.1 as the generator. The generator places the fixed inventory of each task at native scale, using the placement stage of Plan--Retrieve--Place (Appendix~\ref{app:generation-protocols}). The initial placement (R0) fixes instance identities and counts; revisions change only positions and yaw rotations, and every response contains all instances.

Each scene has one R0 layout, shared by three branches. In every round, the generator receives the task, the frozen object plan and asset selection, the previous layout, and a branch-specific evaluation report of that layout:
\begin{itemize}
  \item \textbf{No feedback.} The report is empty, and the generator is asked to inspect the layout itself.
  \item \textbf{Scores.} The report contains the room score and the scores of the five metrics that enter the \frozenassets{} overall score: Collision, Support, Out-of-bounds, Function, and Placement.
  \item \textbf{Report.} The report contains the same scores and every confirmed defect of the previous layout. Each defect record gives the metric, the check type, the affected objects, a severity, the evaluator's reason, and the measurements attached to the check (Table~\ref{tab:app-feedback-record}). Unresolved items are listed separately as neither defects nor passes.
\end{itemize}
The branches share one system prompt. For the Report branch, it adds that the report lists confirmed defects with their evidence and omits passing checks, that unlisted objects and relations are not certified as correct, and that the generator should correct the confirmed defects while preserving the rest of the layout where possible. In all branches, the report is supplied as data and cannot override the placement contract.

\paragraph{Pipeline scoring protocol.}
For this pipeline-based refinement study, we report the evaluator's
recorded benchmark score on a 0--1 scale, retaining its original
metric scores and layer aggregation. The motivation for using a
separate protocol is to preserve discrimination among low-quality
layouts: large accumulated defect penalties can compress scores
near the lower bound and obscure differences between revisions.
The reported score is
\[
  s_{\mathrm{pipeline}}=0.3s_{\mathrm{physical}}
                        +0.7s_{\mathrm{quality}},
\]
where the two terms are the physical and scene-quality layer scores
recorded by the evaluator. The scene-quality term retains the
original evaluator's metric contributions, excluding its visual
metrics. All three branches use the same reported score definition.
The feedback delivered to the Scores and Report branches contains
only the five retained \frozenassets{} metrics and their
renormalized total, so it differs from the full benchmark score
reported here. This study's reporting protocol differs from the
main 0--100 GREP protocol in Appendix~\ref{app:scoring}; its scores
are not obtained by dividing those scores by 100.
Each branch ran three revisions, therefore we analyze R1--R3.

\begin{table}[h]
  \centering
  \small
  \caption{A defect record from the diagnostic report\textsuperscript{*} on the R0 layout of S104, shortened to the fields listed. Records also carry the verdict, a defect category, the check scope, an attribution field, and check identifiers, omitted here.}
  \label{tab:app-feedback-record}
  \newcommand{\appcell}{\raggedright\hspace{0pt}}%
  \begin{tabular}{@{}lp{9.6cm}@{}}
    \toprule
    Field & Content \tabularnewline
    \midrule
    Metric & \appcell Function \tabularnewline
    Check type & \appcell Relative-use geometry \tabularnewline
    Objects & \appcell Piano bench, with the digital piano \tabularnewline
    Severity & \appcell Blocked \tabularnewline
    Reason & \appcell ``The bench is positioned about 2.0\,m from the piano rather than within seated operating reach, so it cannot function as seating for the piano in the authored arrangement.'' \tabularnewline
    Measurements & \appcell Center distance 2.0\,m; the seating side of the bench and the control side of the piano each at $90^\circ$ to the direction of the other object \tabularnewline
    \bottomrule
  \end{tabular}
\end{table}

\subsection{Per-Scene Results}
\label{app:feedback-results}

Table~\ref{tab:app-feedback-scenes} lists every score under the
pipeline scoring protocol. Averaged over scenes and R1--R3,
No feedback scores 0.552, Scores 0.565, and Report 0.626,
from a shared R0 mean of 0.585. Paired by scene and round
(15 pairs), Scores differs from No feedback by $+0.013$
(standard error 0.012), Report by $+0.074$
(standard error 0.027; positive in 11 pairs), and Report
from Scores by $+0.061$ (standard error 0.025; positive in 12 pairs).
These pairs contain repeated observations from the same five scenes.
The Report branch has the highest mean in four of the five scenes.
S103 is the exception: its Scores branch, which never changed the
layout, has the highest mean.

\begin{table}[h]
  \centering
  \small
  \caption{Pipeline scores of the refinement study (0--1). R0 is shared by the three branches of a scene. A dagger marks a revision whose layout is identical to that of the previous round.}
  \label{tab:app-feedback-scenes}
  \begin{tabular}{@{}llrrrrr@{}}
    \toprule
    Scene & Branch & R0 & R1 & R2 & R3 & Mean R1--R3 \\
    \midrule
    S101 & No feedback & 0.534 & 0.489 & 0.456$^\dagger$ & 0.526 & 0.490 \\
     & Scores &  & 0.494 & 0.517 & 0.499 & 0.503 \\
     & Report &  & 0.435 & 0.526 & 0.670 & 0.544 \\
    \addlinespace
    S102 & No feedback & 0.608 & 0.542$^\dagger$ & 0.458$^\dagger$ & 0.403$^\dagger$ & 0.468 \\
     & Scores &  & 0.468$^\dagger$ & 0.493$^\dagger$ & 0.458$^\dagger$ & 0.473 \\
     & Report &  & 0.569 & 0.612 & 0.712 & 0.631 \\
    \addlinespace
    S103 & No feedback & 0.525 & 0.566 & 0.618$^\dagger$ & 0.628$^\dagger$ & 0.604 \\
     & Scores &  & 0.591$^\dagger$ & 0.659$^\dagger$ & 0.615$^\dagger$ & 0.622 \\
     & Report &  & 0.656 & 0.508 & 0.590 & 0.585 \\
    \addlinespace
    S104 & No feedback & 0.605 & 0.627$^\dagger$ & 0.620 & 0.594 & 0.613 \\
     & Scores &  & 0.615$^\dagger$ & 0.615$^\dagger$ & 0.629$^\dagger$ & 0.619 \\
     & Report &  & 0.665 & 0.616 & 0.706 & 0.662 \\
    \addlinespace
    S105 & No feedback & 0.653 & 0.527$^\dagger$ & 0.665 & 0.563$^\dagger$ & 0.585 \\
     & Scores &  & 0.610 & 0.598 & 0.616$^\dagger$ & 0.608 \\
     & Report &  & 0.715 & 0.713 & 0.702 & 0.710 \\
    \bottomrule
  \end{tabular}
\end{table}

\subsection{Layout Changes}
\label{app:feedback-changes}

A revision counts as changed when the hash of its placement file differs from that of the previous round; R1 is compared with R0. The layout changed in 6 of 15 revisions without feedback, 5 of 15 with scores, and 15 of 15 with reports. The score branch returned the R0 layout in every round for S102, S103, and S104, and the no-feedback branch did so for S102. Unchanged revisions are evaluated again rather than copied, so their scores differ from R0 only through evaluation. For S103, the three unchanged layouts of the score branch score 0.591 to 0.659 against 0.525 for R0, because the evaluation of R0 confirmed one Placement defect that the later evaluations did not.

\subsection{Example: Function in S104}
\label{app:feedback-case}

S104 asks for a media, music, game, and recreation room of 10.4 by 7.6\,m with 30 objects. The evaluation of its R0 layout confirmed 12 Function defects. Eight objects, among them the television, soundbar, media console, both studio monitors, and the guitar amplifier, had their functional side facing the room boundary. Three access regions were blocked, for example the access face of a bookcase 0.22\,m from the boundary. The piano bench stood 2.0\,m from the digital piano (Table~\ref{tab:app-feedback-record}). The Report branch received these records together with the defects of the other metrics.

In R1, none of the eleven orientation and clearance defects was confirmed again, and the recorded Function score rose from 0.447 to 0.750, remaining at 0.750 in R2. Over the same rounds, Function stayed between 0.400 and 0.480 in the other two branches, whose evaluations confirmed 11 to 13 Function defects in every round. The revision was incomplete. The piano bench moved next to the piano but across its control side, with its back toward the keyboard, and was reported in R1, R2, and R3 without being resolved. R1 also introduced a new defect, a soundbar placed at the position of the television, which was cleared in R2. In R3, the recorded Function score remained 0.750 despite two further defects: the television and the sofa were each turned $63^\circ$ away from the other, so viewing required turning the head, and the soundbar blocked the front of the media console. The reported pipeline score was 0.605 at R0, followed by 0.665, 0.616, and 0.706. Physical defects remained: ten Collision defects were recorded in every round, and Support defects rose from two in R0 to five in R1 and R2.

\subsection{Evaluation Records}
\label{app:feedback-records}

The study comprises 50 evaluations: five R0 layouts and 45 revisions. Every retained evaluation is complete with full judgment coverage. Nine were retained from a second or third attempt after the earlier attempts did not produce a complete evaluation; incomplete attempts were discarded and were neither scored nor fed back. The feedback for each round was produced from the retained evaluation of the previous layout.

\end{document}